\documentclass[12pt]{article}

\usepackage{amsmath,amssymb,amsthm}
\usepackage[colorlinks=true,linkcolor=blue,citecolor=blue,urlcolor=blue]{hyperref}

\newtheorem{theorem}{Theorem}[section]
\newtheorem{lemma}[theorem]{Lemma}
\newtheorem{definition}[theorem]{Definition}
\newtheorem{remark}[theorem]{Remark}

\numberwithin{equation}{section}

\title{A Refinement of Vapnik--Chervonenkis' Theorem}
\date{}

\begin{document}

\begin{titlepage}
\thispagestyle{empty}

\begin{center}
{\LARGE\bfseries A Refinement of Vapnik--Chervonenkis' Theorem\par}
\vspace{1.25cm}

\begin{minipage}[t]{0.95\textwidth}
\begin{center}
{\large Alex Iosevich\footnotemark[1]\par}
\vspace{0.2cm}
Department of Mathematics, University of Rochester\\
Rochester, NY 14627, USA
\end{center}
\end{minipage}

\vspace{0.75cm}

\begin{minipage}[t]{0.95\textwidth}
\begin{center}
{\large Armen Vagharshakyan\footnotemark[2]\par}
\vspace{0.2cm}
Institute of Mathematics at NAS RA\\
Yerevan, Armenia\\
Yerevan State University\\
Yerevan, Armenia\\[0.2cm]
Corresponding author: \texttt{avaghars@kent.edu}
\end{center}
\end{minipage}

\vspace{0.75cm}

\begin{minipage}[t]{0.95\textwidth}
\begin{center}
{\large Emmett Wyman\footnotemark[3]\par}
\vspace{0.2cm}
Department of Mathematics, Binghamton University\\
Binghamton, NY 13902, USA
\end{center}
\end{minipage}

\end{center}

\vfill

\footnotetext[1]{A. Iosevich was supported in part by the National Science Foundation under NSF DMS--2154232.}
\footnotetext[2]{A. Vagharshakyan was supported by the Higher Education and Science Committee of RA (Research Project No 24RL-1A028).}
\footnotetext[3]{E. Wyman was supported in part by the Natural Sciences and Engineering Research Council of Canada, NSERC (GR030571 and GR030540).}

\end{titlepage}

\begin{abstract}
Vapnik--Chervonenkis' theorem is a seminal result in machine learning. It establishes sufficient conditions for empirical probabilities to converge to theoretical probabilities, uniformly over families of events. It also provides an estimate for the rate of such uniform convergence.

We revisit the probabilistic component of the classical argument. Instead of applying Hoeffding's inequality at the final step, we use a normal approximation with explicit Berry--Esseen error control. This yields a moderate-deviation sharpening of the usual VC estimate, with an additional factor of order $(\varepsilon\sqrt{n})^{-1}$ in the leading exponential term when $\varepsilon\sqrt{n}$ is large.


\end{abstract}

\section{Introduction}

Vapnik--Chervonenkis' theorem (also referred to as the Fundamental theorem of statistical learning) is a seminal result in machine learning. As such, it is routinely cited in textbooks on machine learning (see for example \cite{Wang} and \cite{DuZhang}) and has been a subject of philosophical discussions (see for example \cite{SpeldaStritecky}).

The theorem establishes sufficient conditions for empirical probabilities to converge to theoretical probabilities, uniformly over families of events. Moreover, by combining combinatorial arguments with exponential concentration inequalities, the original proof provides an estimate for the rate of such uniform convergence.

At a structural level, the classical proof separates into two components: a combinatorial reduction governed by the growth function of the class, and a probabilistic estimate controlling deviations for a fixed set via an exponential tail bound. The present work leaves the combinatorial mechanism unchanged and focuses exclusively on refining the final probabilistic step.

In this paper, we revisit that probabilistic component. Rather than relying on Hoeffding's inequality, we use a normal approximation with explicit Berry--Esseen error control, leading to improved constants in moderate deviation regimes while preserving the standard VC framework.

Let $(X,\mathcal{B}(X),P)$ be a probability space and let $x_1,\dots,x_n$ be independent and identically distributed random variables taking values in $X.$ For a measurable set $A\in\mathcal{B}(X),$ its empirical probability $d_n(A)$ is defined by
$$
d_n(A)=\frac{1}{n}\sum_{i=1}^n \mathbf{1}_{\{x_i\in A\}}.
$$

Bernoulli's theorem, being a special case of the law of large numbers, ensures convergence of empirical probabilities to the theoretical probability for a fixed measurable set $A$.

\begin{theorem}[Bernoulli]\label{thm:bernoulli}
For every $\varepsilon>0$ and measurable $A\in \mathcal{B}(X)$,
$$
P\left(|d_n(A)-P(A)|>\varepsilon\right)\to 0
$$
as $n\to\infty$.
\end{theorem}

The Vapnik--Chervonenkis theorem characterizes when the convergence in Theorem \ref{thm:bernoulli} is uniform over a family $S.$ We write
\begin{equation}\label{bns}
B_n(S)=P\left(\sup_{A\in S}|d_n(A)-P(A)|>\varepsilon\right).
\end{equation}

\begin{theorem}[Vapnik--Chervonenkis]\label{thm:vc}
Let $S\subset\mathcal{B}(X)$ be a family of measurable sets. Then for $\varepsilon>0$,
\begin{equation}\label{eq:vc-hoeffding}
B_n(S)\le 2m_n(S)e^{-2n\varepsilon^2},
\end{equation}
where $m_r(S)$ is the growth function of the family $S$ defined by
$$
m_r(S)=\max_{\{x_1,\dots,x_r\}\subset X}\#
\left\{
\{x_1,\dots,x_r\}\cap A:A\in S
\right\}.
$$
\end{theorem}

We improve Theorem \ref{thm:vc} in the moderate deviation regime by replacing the final Hoeffding step with a normal approximation.

\begin{theorem}\label{thm:main}
Let $S\subset\mathcal{B}(X)$ and let $m_n(S)$ be its growth function. Then for $\varepsilon>0$,
\begin{equation}\label{eq:main}
B_n(S)\le m_n(S)\left(\frac{e^{-2n\varepsilon^2}}{\varepsilon\sqrt{2\pi n}}+\frac{C}{\sqrt{n}}\right),
\end{equation}
where $C$ is a Berry--Esseen constant. In particular, one may take
\begin{equation}\label{eq:BE-constant}
C\le 0.4748.
\end{equation}
\end{theorem}

\begin{remark}
If $S$ has finite VC dimension $d$, the Sauer--Shelah lemma yields
$$
m_n(S)\le \left(\frac{en}{d}\right)^d,
$$
producing an explicit polynomial--exponential bound when combined with \eqref{eq:main}.
\end{remark}

\section{Preliminaries}

\subsection{Growth function and VC dimension}

\begin{definition}[VC dimension]
The VC dimension of $S$ is the largest integer $d$ such that $m_d(S)=2^d$.
\end{definition}

\subsection{Normal approximation tools}

The best result known so far for Berry--Esseen error control has been proved by Shevtsova \cite{Shevtsova}.

\begin{theorem}[Berry--Esseen for Bernoulli]\label{thm:be-bern}
Let $\xi_1,\dots,\xi_n$ be independent and identically distributed Bernoulli$(p)$ variables with variance $\sigma^2=p(1-p)$. Let $F_n$ be the distribution function of
$$
\frac{\sum_{i=1}^n\xi_i-np}{\sigma\sqrt{n}},
$$
and let $\Phi$ denote the cumulative distribution function of the standard normal distribution. Then
$$
\sup_t |F_n(t)-\Phi(t)|\le \frac{C}{\sqrt{n}},
$$
where $C$ may be taken as in \eqref{eq:BE-constant}.
\end{theorem}

We complement Theorem \ref{thm:be-bern} with the following upper-tail inequality for the standard normal distribution.

\begin{lemma}[Mill's ratio]\label{lem:mills}
For $x>0$,
$$
1-\Phi(x)\le \frac{e^{-x^2/2}}{x\sqrt{2\pi}}.
$$
\end{lemma}

\section{Proof of Theorem \ref{thm:main}}

Note that for a fixed outcome $\omega$ in the underlying probability space $X,$ the supremum in \eqref{bns} depends only on the trace of $A$ on the set $\{x_1(\omega),\dots,x_n(\omega)\}$. Correspondingly, define $S'(\omega)\subset S$ to be a maximal subfamily of $S$ whose intersections with the set $\{x_1(\omega),\dots,x_n(\omega)\}$ are distinct. Consequently, $\#S'(\omega)\le m_n(S),$ and
$$
\sup_{A\in S}|d_n(A)-P(A)|=\sup_{A\in S'(\omega)}|d_n(A)-P(A)|.
$$
By the union bound,
$$
B_n(S)\le m_n(S)\cdot \sup_{A\in S}P\left(\left|d_n(A)-P(A)\right|>\varepsilon\right).
$$

Fix $A\in S$, set $p=P(A)$, and let $\xi_i=\mathbf{1}_{\{x_i\in A\}}$. Then
$$
d_n(A)=\frac{1}{n}\sum_{i=1}^n \xi_i.
$$
Write $\sigma^2=p(1-p)$. The event $|d_n(A)-p|>\varepsilon$ is the same as
$$
\left|\sum_{i=1}^n\xi_i-np\right|>\varepsilon n.
$$
Standardizing,
$$
P\left(|d_n(A)-p|>\varepsilon\right)=P\left(\left|\frac{\sum_{i=1}^n\xi_i-np}{\sigma\sqrt{n}}\right|>\frac{\varepsilon\sqrt{n}}{\sigma}\right).
$$
By Theorem \ref{thm:be-bern} and symmetry of $\Phi$,
$$
P\left(|d_n(A)-p|>\varepsilon\right)\le 2\left(1-\Phi\left(\frac{\varepsilon\sqrt{n}}{\sigma}\right)\right)+\frac{C}{\sqrt{n}}.
$$
Applying Lemma \ref{lem:mills} with $x=\varepsilon\sqrt{n}/\sigma$ yields
$$
2\left(1-\Phi\left(\frac{\varepsilon\sqrt{n}}{\sigma}\right)\right)\le \frac{2\sigma}{\varepsilon\sqrt{n}}\frac{e^{-\frac{1}{2}\left(\frac{\varepsilon\sqrt{n}}{\sigma}\right)^2}}{\sqrt{2\pi}}.
$$
Using $\sigma^2\le 1/4$ gives $\sigma\le 1/2$ and
$$
\frac{1}{2}\left(\frac{\varepsilon\sqrt{n}}{\sigma}\right)^2\ge 2n\varepsilon^2.
$$
Combining these inequalities yields
$$
P\left(|d_n(A)-p|>\varepsilon\right)\le \frac{e^{-2n\varepsilon^2}}{\varepsilon\sqrt{2\pi n}}+\frac{C}{\sqrt{n}}.
$$
Substituting into the union bound completes the proof of \eqref{eq:main}.

\section{Comparison with classical bounds}

Note that Hoeffding's inequality \cite{Hoeffding} yields \eqref{eq:vc-hoeffding}.

The point of \eqref{eq:main} is that, for moderate deviations, the normal tail has an additional factor of order $(\varepsilon\sqrt{n})^{-1}$ which is not seen by Hoeffding.

To make the gain explicit, compare the leading exponential terms in \eqref{eq:vc-hoeffding} and \eqref{eq:main}. Whenever $\varepsilon\sqrt{n}$ is large enough that the Berry--Esseen error term $C/\sqrt{n}$ is dominated by the normal-tail contribution, the prefactor in \eqref{eq:main} is strictly smaller than the prefactor $2$ in \eqref{eq:vc-hoeffding}. In particular, if $\varepsilon\sqrt{n}\to\infty$ then
$$
\frac{C}{\sqrt{n}}=o\left(\frac{e^{-2n\varepsilon^2}}{\varepsilon\sqrt{n}}\right),
$$
and the normal-tail term governs the bound.

On the other hand, if $\varepsilon\sqrt{n}$ stays bounded, then the term $C/\sqrt{n}$ may be comparable to or larger than the exponential term. In that small-deviation regime, the present argument does not claim an improvement over the standard VC inequality. Thus the gain is a constant-level sharpening in the moderate deviation range, rather than a change in the underlying VC mechanism governed by the growth function.

\section{Examples}

\subsection{Intervals in $\mathbb{R}$}

For intervals in $\mathbb{R}$, the VC dimension equals $2$ and one has $m_n(S)\le n^2$.

\subsection{Half-spaces in $\mathbb{R}^d$}

For half-spaces in $\mathbb{R}^d$, the VC dimension equals $d+1$ and one has
$$
m_n(S)\le \left(\frac{en}{d+1}\right)^{d+1}.
$$

\section{Discussion and open problems}

The bound \eqref{eq:main} sharpens constants in VC-type estimates using a normal approximation. In the moderate deviation regime, the factor $(\varepsilon\sqrt{n})^{-1}$ appearing in the normal tail is consistent with classical asymptotics for sums of independent Bernoulli variables and is therefore expected to be of the correct order in general.

Natural follow-up questions include optimal switching between \eqref{eq:vc-hoeffding} and \eqref{eq:main}, extensions to relative deviations, and refinements based on sharper normal approximations.

\end{document}